\pdfoutput=1

\documentclass[11pt]{article}

\usepackage[final]{acl}
\usepackage{times}
\usepackage{latexsym}

\usepackage[T1]{fontenc}
\usepackage{array}
\usepackage{booktabs}
\usepackage{multirow}
\usepackage{tabularx}
\usepackage{makecell}
\usepackage{hyperref}

\usepackage[utf8]{inputenc}

\usepackage{microtype}

\usepackage{inconsolata}

\usepackage{graphicx}

\title{Improve  Decoding Factuality by Token-wise Cross Layer Entropy of Large Language Models}

\author{
    Jialiang Wu\textsuperscript{1}\thanks{Work done during an internship at Beijing Wispirit Tech.},
    Yi Shen\textsuperscript{2},
    Sijia Liu\textsuperscript{1},
    Yi Tang\textsuperscript{3},
    Sen Song\textsuperscript{4},
    Xiaoyi Wang\textsuperscript{2},
    Longjun Cai\textsuperscript{2,4}\thanks{Corresponding author} \\
     \textsuperscript{1}Harbin Institute of Technology, \textsuperscript{2}Beijing Wispirit Technology,\\
    \textsuperscript{3}Xuanwu Hospital, \textsuperscript{4}Tsinghua University \\
        \texttt{\{jialiang.cg, owen.shen.1988, cailongjun\}@gmail.com} \\
        \texttt{sijia.liu@stu.hit.edu.cn, tangyi@cibr.ac.cn} \\
        \texttt{songsen@tsinghua.edu.cn, wangxiaoyi@66nao.com}
}

\begin{document}
\maketitle
\begin{abstract}

Despite their impressive capacities, Large language models (LLMs) often struggle with the hallucination issue of generating inaccurate or fabricated content even when they possess correct knowledge.
In this paper, we extend the exploration of the correlation between hidden-state prediction changes and output factuality into a deeper, token-wise level. 
Based on the insights , we propose \textit{cross-layer \textbf{E}ntropy e\textbf{N}hanced \textbf{D}ecoding \textbf{(END)}}, a decoding method that mitigates hallucinations without requiring extra training.
END leverages inner probability changes across layers to individually quantify the factual knowledge required for each candidate token,
and adjusts the final predicting distribution to prioritize tokens with higher factuality.
Experiments on both hallucination and QA benchmarks demonstrate that END significantly enhances the truthfulness and informativeness of generated content while maintaining robust QA accuracy. 
Moreover, our work provides a deeper perspective on understanding the correlations between inherent knowledge and output factuality.\footnote{The source code is available at \url{https://github.com/Arcade-Master/END}.}


\end{abstract}

\section{Introduction}

Large language models (LLMs) have demonstrated remarkable capabilities across numerous natural language processing (NLP) applications \cite{zhao2023survey,openai2024gpt4technicalreport}. Despite their impressive performance, the issue of generating fabricated content, commonly referred to as "hallucinations" \cite{ji2023halllusurvey1,zhang2023hallusurvey2}, remains a persistent challenge for LLMs.
This problem hinders the broader application of LLMs in industries that highly demand trustworthiness and accuracy.

\begin{figure}[t]
  \includegraphics[width=1.0\columnwidth]{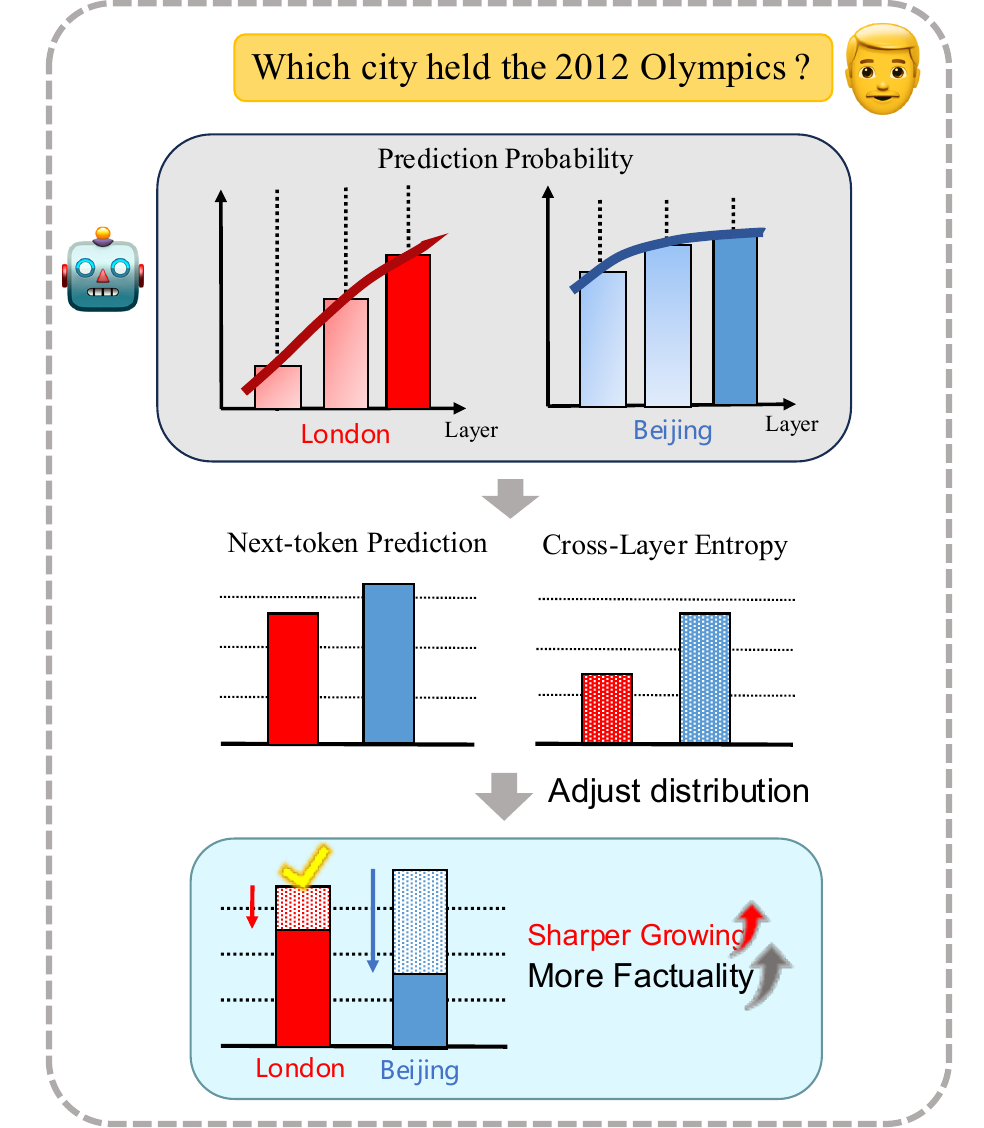}
  \caption{Illustration of our proposed method for improving LLMs' decoding factuality. The prediction probability of the wrong token `Beijing' is adjusted and surpassed, since the correct token `London' requires more factual knowledge during inference. }
  \label{fig:head}
\end{figure}
Recently, various methods have been proposed to mitigate hallucinations, including training with high-quality data, aligning with de-hallucination pair, and integrating external knowledge sources. 
However, these methods often involve high computational costs or demands knowledge bases, which may not be accessible in many application scenarios.
Also, some studies \cite{wei2022chainCOTEVEN,saunders2022selfEVEN} have found that even when LLMs possess the corresponding knowledge, they can still be susceptible to generating hallucinations.
To tackle this issue, recent research has focused on the internal representations of models, exploring the correlation between hidden states and output truthfulness. 
Especially,
\citet{chuang2023dola} discovered that the prediction distribution of LLMs remains actively fluctuating in higher layers when generating factual tokens, while it remains almost unchanged when producing other easy tokens. 
Similarly, 
\citet{chen2024activation} reveals that the correct generation typically exhibits sharper context activation within the inner layers across in-context tokens.
Furthermore, we extend the investigation from step-level to token-level, analyzing the hidden-state change across layers for each token within a generation step.
Our findings reveal that tokens, associated with factual knowledge and answer correctness, exhibit a sharp growing trend of predicting probability with notable changes in the higher layers.
This aligns with previous findings and provides a more granular explanation at the token level.


To this end, we propose cross-layer entropy enhanced decoding (END), a novel decoding method that leverages the change of cross-layer predictions to amplify the emerging of factual knowledge.
As shown in Fig.\ref{fig:head},
instead of selecting a certain caliber layer, our method processes the overall growing trend across model layers, offering a more reliable quantification of factual knowledge for each candidate token individually.
Without extra training required, END could be directly applied to LLMs and effectively improve generation factuality.

We evaluate our method on both hallucination benchmarks (TruthfulQA and FACTOR) and general QA benchmarks (TriviaQA and Natural Questions). 
Experimental results demonstrate that our proposed END significantly enhances the factuality of model generation while maintaining robust basic QA performance. 
Also, we further extend the experiments to various LLM backbones of different scales and types, verifying its generalizability of application.
Overall, our work not only introduces an effective decoding method to enhance generation factuality, but also provides a new perspective on exploring correlation between inner hidden states and output truthfulness at a token level.

\section{Related Work}
Recently, various methods have been proposed to improve LLM's generating factuality to mitigate hallucinations. These include, but are not limited to, supervised fine-tuning with high-quality data \cite{tian2023fineSFT,zhou2024limaSFT}, reinforcement learning with truthful preference pairs \cite{sun2023aligningRL,yang2023alignmentRL}, retrieval-augmented generation that integrates external knowledge \cite{chern2023factoolRAG},and editing knowledge-related inner representations or parameter-efficient modules \cite{zhang2024truthx,hu2024separateLORA}.

Our research focuses on the field of constraint decoding, which involves applying intervention strategies during model's generation process.
Notably, Inference-Time Intervention (ITI) \cite{li2024inferenceITI} employs probes to locate truthfulness signals within attention heads, while Repe \cite{zou2023representationZOU} locates those within critical layers, then editing on the direction of truthfulness to modify model decoding.
Contrast Decoding (CD) \cite{li2022contrastiveCD} and later Induced-then-Contrast Decoding (ICD) \cite{zhang2023alleviatingICD} contrasts logits from an expert model against those from a weak model, amplifying the knowledge reflected in their differences.
Activation Decoding \cite{chen2024activation} leverages the correlation between context activation sharpness and answer correctness,
incorporating in-context entropy into decoding to improve factuality.

The most relevant work to ours is  DoLA \cite{chuang2023dola}, which selects a single, most distinct layer to contrast with the final layer, amplifying the factual knowledge boosted in higher layers.
However, the change of inner predictions varies by candidate tokens, which means that, at a given generation step, factual tokens may exhibit different growing trends. Therefore, selecting a single caliber layer for all tokens is not accurate and can lead to false negative and false positive problems. 
Unlikely, we propose to process the prediction changes across layers individually for each token. By quantifying their growing trend, we can leverage internal information more accurately to enhance the factuality of generation.


\section{Empirical Findings}

\begin{figure*}[!t]
  \includegraphics[width=1\textwidth]{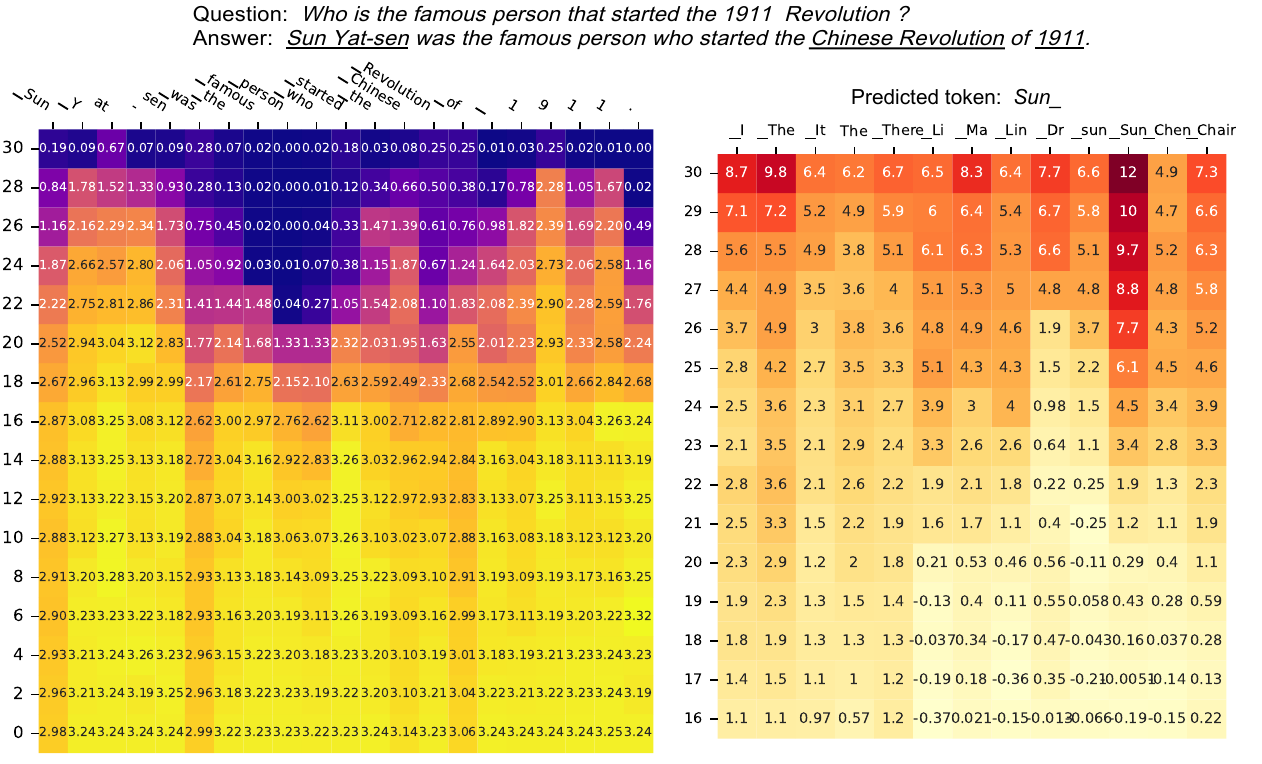}
  \caption{\textbf{(1)} The left figure illustrates the predicting distribution differences, measure by KL-divergence, between the final layer and even-numbered early layers of the whole output sentence. Row names are indices of the early layer used for contrasting and column names are decoded tokens at each generation step. \textbf{(2)} The right figure illustrates the predicting probabilities of high probability candidate tokens among higher layers at the first generation step of token `$\_Sun$'. Row names are indices of the layer and  column names are candidate tokens.}
  \label{fig:finding}
\end{figure*}

Previous works \cite{chuang2023dola,halawi2023overthinking,schuster2022confident} discovered that, when generating tokens that require factual knowledge, such as name entities, dates and locations, 
model tends to be still changing its predictions in the last few layers since it is potentially injecting more factual knowledge into inference.
Contrarily, prediction changes are minimal from the middle layers onward when generating `easy' tokens, such as syntactic or functional tokens.
This may be because model has already decided the token to generate at middle and keeps the prediction almost unchanged in afterwards higher layers.
Later work \cite{chen2024activation} also digs into hidden states and finds that, successful activation with sharp in-context logits indicates higher chance of answer correctness.


To motivate our approach, we conduct analysis into hidden-state predictions among layers with LLaMA-2-7B model (32 layer).
Following previous studies, we first use KL-divergence to measure the prediction differences between inner layers and the final layer, and observe the phenomenon in Fig.\ref{fig:finding} (Left). 
The generation steps of factual tokens like \textit{`Sun-Yat sen'}, \textit{`Chinese Revolution'} and \textit{`1911'} present active changes in higher layers while others tend to be stable.
Then, we further dig into vocabulary-level to analyze the prediction change of each candidate token in a generation step.
As shown in Fig.\ref{fig:finding} (Right), among all candidates in the model vocabulary, the prediction probability of factuality tokens, i.e. name entities \textit{(Li, Ma, Lin, Dr, Sun, Chen)}, tend to grow sharply from a relatively low value in higher layers while others' tend to grow in a relatively gentle trend.

Therefore, it could be concluded that, it is the factual candidate tokens that play a dominant role in driving the changes in the prediction distribution in higher layers at the step of generating factual token, which explains the phenomenon observed by previous works. 
Also, for factual candidate tokens, the position and trend of their prediction changes across layers are not the same, which makes it impossible to select one best caliber layer to capture the knowledge emerging for all.
Based on these, it is natural to use token level cross-layer information instead of the whole voabulary-level prediction change to amplify factuality.
In this way, we propose to individually capture the prediction change of each candidate token so as to better quantify the factual knowledge required and use this information to help enhance model decoding.

\section{Methodology }

\begin{figure*}[t]
  \includegraphics[width=1\textwidth]{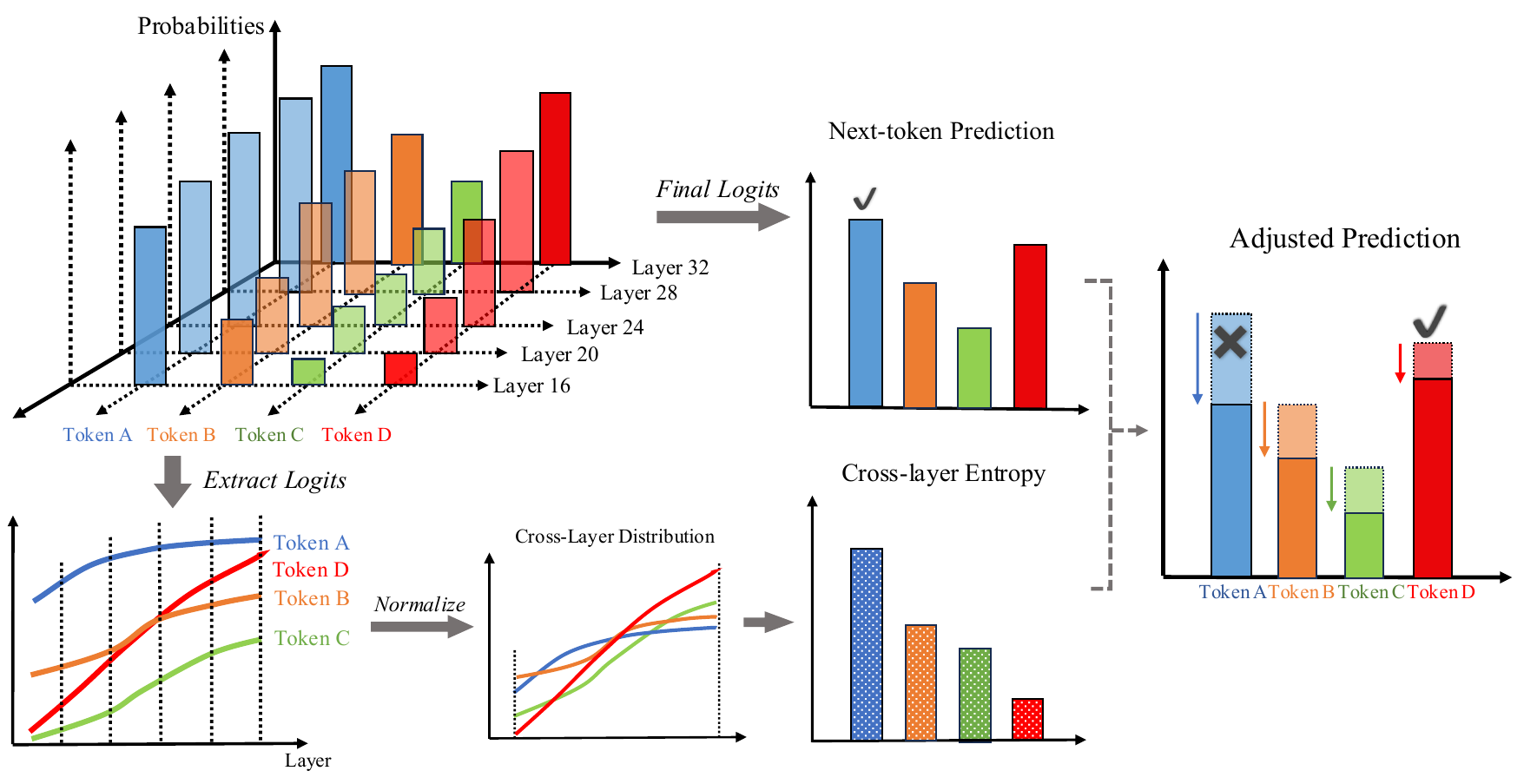}
  \caption{Workflow of our proposed method. While the predicting probability of token A remains almost unchanged in higher layers, that of token D shows a sharp growing trend from a low value to the second-highest. The cross-layer entropy adjusts the final next-token prediction by suppressing the token A for low factuality and highlighting token D for both high prediction probability and high factuality.}
  \label{fig:workflow}
\end{figure*}

Based on these findings, we propose END, a decoding enhancement method that
can be directly applied to mitigate hallucinations without incurring additional training costs.
As illustrated in Fig.\ref{fig:workflow}, our method measures the internal prediction changes of candidate tokens, introduces cross-layer entropy to quantify the factual knowledge of inference, and adjusts model's next-token prediction by favoring factuality token to improve the informativeness and truthfulness of the generated content.

\subsection{Cross-Layer Entropy}

Large language models typically consist of $N$ stacked transformer layers, followed by an affine layer that maps the internal representations to a next-token probability distribution.
We denote the hidden state of the $l$-th layer as $h^{(l)}$ , the classification head as $\varphi(\cdot)$, and $v_t$ as generation token at step $t$ over the vocabulary set $V$. The prediction probability from the $l$-th layer can be expressed as: 
\begin{equation}
  \label{eqn:0}
P_l(v_t \mid v_{1:t-1}) = softmax(\varphi(h_{t}^{(l)})), v_t \in V
\end{equation}

Here, the prediction probability $p_l(v_t)$  is a $k$-dimension vector that includes the $l$-th layer's prediction values for all $k$ candidate tokens in the model vocabulary.

For each candidate token, we extract its prediction values of higher layers and use them to constitute a cross-layer probability distribution $D$, which characterizes the token's prediction change across layers, reflecting its growing trend.
The probability used to build cross-layer distribution is calculated as Equation \ref{q}, where $Layer$ represents the set of layers ranging from middle to high.
\begin{equation}
  \label{q}
q_l(v_t) = \frac{P_l(v_t)}{ {\textstyle \sum_{i \in Layer}} P_i(v_t)} 
\end{equation}

We normalize the prediction values to bring them into a consistent range so that the trend of prediction change could be directly compared across candidate tokens regardless of differences in absolute probability values.

As the finding suggests, functional tokens with factual knowledge injected during inference often exhibit sudden growing with unstable changes.
As a result,  their cross-layer distributions are likely to present a sharp or highly volatile trend.
In contrast,
for other easy or unrelated tokens,
the prediction probabilities grow slightly or remain relatively unchanged in higher layers, leading to flatter and more stable distributions.

Therefore, to quantify the factual knowledge required of each candidate token, we introduce \textbf{cross-layer entropy} to measure the growing trend within the cross-layer probability distribution:
\begin{equation}
  \label{entropy}
Entropy(v_t) = \sum_{l \in Layer}q_l(v_t)\log{q_l(v_t)}
\end{equation}
A low cross-layer entropy value represents a sharp predicting distribution, indicating that the candidate token is more closely associated with factual knowledge and answer correctness.

\begin{table*}[!t]
  \centering
  \renewcommand{\arraystretch}{1}
  \setlength\tabcolsep{8pt}%
  \begin{tabular}{c cccc ccc}
    \toprule
       \multirow{2}{*}{\textbf{Method}}  & \multicolumn{4}{c}{\textbf{Open-ended Generation}} & \multicolumn{3}{c}{\textbf{Multiple-Choice}} \\
        \cmidrule(lr){2-5} \cmidrule(lr){6-8}
        & \%Truth & \%Info & \%Truth$\ast$Info & \%Reject & MC1 & MC2 & MC3 \\
    \midrule
    Greedy decoding    & 58.36 & 80.54 & 39.41 & 20.32 & 33.5 & 50.6 & 24.4  \\
    DoLa     & 61.93 & 86.54 & 48.96 & 13.95 & 33.7 & 50.5 & 24.6  \\
    Activation Decoding    & 55.08 & 90.45 & 46.02 & 9.67 & 33.0 & \textbf{51.4} & \textbf{25.2}  \\
    Our method    & \textbf{66.71} & \textbf{94.12} & \textbf{61.20} & \textbf{5.51} & \textbf{34.4} & \textbf{51.4} & \textbf{25.2}  \\
    \bottomrule
  \end{tabular}
  \caption{
    Main result on TruthfulQA open-ended generation task and multiple-choice task. Best performance of each metric is highlighted in \textbf{bold}. The scores of multiple-choice task is obtained from previous authorized work \cite{chen2024activation} and those of open-ended generation are re-evaluated by replicating because of the change of GPT-3 model version.
  }
  \label{tab: main}
\end{table*}

\subsection{Factuality Enhanced Decoding }

To improve generation quality and mitigate hallucinations, tokens associated with factual knowledge should be amplified during decoding while unrelated ones should be suppressed.
We implement this by using cross-layer entropy to adjust the next-token prediction from the final layer:
\begin{equation}
  \label{final}
P_{Final}(v_t) = e^{-\lambda Entropy(v_t)}P_N(v_t)
\end{equation}
where $\lambda $ is a hyperparameter controlling the influence of cross-layer entropy and $N$ is the index of final layer.

Additionally, following the approach of  \citet{li2022contrastiveCD}, we also introduce a filtered subset $V_{head}$  to improve inference efficiency under open-ended generation settings, $\alpha$ is a threshold parameter
\begin{equation}
  \label{subset}
      \resizebox{0.45\textwidth}{!}{$V_{head}(v_t) = \left \{ v_t \in V:P_N(v_t) \ge \alpha \max_w P_N(w)  \right \}$}
\end{equation}
where $\alpha$ is a threshold hyperparameter.
Calculating cross-layer entropy and adjusting probability distribution can lead to  substantial computational cost, especially in open-generation settings. For instance, under LLaMa-series model settings, each generation step involves a  vocabulary size of 32,000 tokens. By filtering out low-probability candidates, we only process a small number of candidates and retain the original logits for others. This approach effectively improves the efficiency of open-ended generation.


\section{Experiment}
\subsection{Setup}

\paragraph*{Datasets}
We consider three types of datasets with various tasks to evaluate our method:

\textbf{TruthfulQA} \cite{lin2021truthfulqa} is the most widely used benchmark for assessing the truthfulness of LLMs. It includes two tasks: multiple-choice and open-ended generation. For multiple choice, the model selects an answer from given options and is evaluated by multiple-choice accuracy (MC1/MC2/MC3). For open-ended generation, the model generates output responses directly, and two fine-tuned GPT-3 models\footnote{Curie model’s fine-tune API is no longer supported since 2024 and we introduce Davinci-002 instead, an enhanced version of GPT-3 model that provides  more precise evaluation.} are introduced to assess truthfulness and informativeness.

\textbf{FACTOR} \cite{muhlgay2023factor} is a reading comprehension benchmark  designed to evaluate a model’s factuality in long-paragraph contexts. It consists of three subsets \textbf{Expert, News} and \textbf{Wiki}, all presented in a multiple-choice format, with performance measured by accuracy.

\textbf{Natural Questions} \cite{kwiatkowski2019natural} and \textbf{TriviaQA} \cite{joshi2017triviaqa}  are  well-established Question Answering benchmarks, evaluated with F1 and Exact Match scores. We include them to assess general QA capabilities.

\paragraph*{Baselines}
We mainly compare our method with light-weight decoding methods that could be directly applied to inference without extra training: 1) \textbf{Greedy deocding}, model's original decoding method that selects the next token with the highest probability; 2) \textbf{DoLa} \cite{chuang2023dola}, that enhances factuality by contrasting logits from inner layers with the final layer; 3) \textbf{Activation decoding} \cite{chen2024activation}, that quantify in-context sharpness to adjust decoding correctness.

\begin{table*}[!t]
  \centering
  \begin{tabular}{l cccc ccc}
    \toprule
        \multirow{2}{*}{\textbf{Method}} & \multicolumn{4}{c}{\textbf{Open-ended Generation}} & \multicolumn{3}{c}{\textbf{Multiple-Choice}} \\
        \cmidrule(lr){2-5} \cmidrule(lr){6-8}
       & \%Truth & \%Info & \%Truth$\ast$Info & \%Reject & MC1 & MC2 & MC3 \\
    \midrule
    LLaMA-2-7B-chat     & 58.36 & 80.54 & 39.41 & 20.32 & 33.5 & 50.6 & 24.4  \\
        \multirow{2}{*}{+ Ours}     & \textbf{66.71}& \textbf{94.12} & \textbf{61.20} & \textbf{5.51} & \textbf{34.4} & \textbf{51.4}& \textbf{25.2} \\
                    &(+8.35) & (+13.58)&(+21.79)&(-14.81)&(+0.9)&(+0.8)&(+0.8)\\
    \midrule
    LLaMA-2-13B-chat     & 60.47&86.54&47.37&13.59&35.3&53.3&26.6  \\
    \multirow{2}{*}{+ Ours}     & \textbf{66.58}& \textbf{96.21}& \textbf{63.04}& \textbf{2.56}& \textbf{35.4}& 53.3& \textbf{26.7}  \\
                & (+6.11) & (+9.67)& (+15.67)& (-11.03)& (+0.1)& -& (+0.1)  \\
    \midrule
    LLaMA-2-70B-chat     & 63.65& 73.93& 37.70& 26.81& 37.3& 56.3& 27.9 \\
    \multirow{2}{*}{+ Ours}     & \textbf{69.16} & \textbf{92.04}& \textbf{62.55}& \textbf{7.22}& \textbf{37.6}& 56.3& \textbf{28.2}  \\
                        &(+5.51) &(+18.11) & (+24.85) & (-19.59)&(+0.3)&-&(+0.3)\\
    \bottomrule
  \end{tabular}
  \caption{
    Experimental result of our method on different scales of LLaMa-2 model on TruthfulQA. The best improved performance for each metric is highlighted in \textbf{bold}. Values in parentheses indicate the improvement over the original greedy decoding. 
  }
  \label{scale}
\end{table*}

\paragraph*{Implementation Details}

We use Llama-2-7B-chat as the backbone model for experiments. Similar to \citet{chuang2023dola}, we also divide all layers into buckets and use the same strategy to select one as $Layer$ set to construct cross-layer distribution. 
The filter threshold $\alpha$ is set to $\left[0.001,0.1\right]$. The entropy adjustment coefficient $\lambda$ is set to $\left[1,3\right]$ for open-ended generation task, and $\left[0.25,0.5\right]$ for multiple choice and QA task. The exact hyperparameter values are determined through validation runs on the respective benchmark.

\subsection{Main Results}

\paragraph*{Results on TruthfulQA}

The experiment results on TruthfulQA are presented in Table \ref{tab: main}.
In the multiple-choice task, our method achieves the highest MC1 score and the equal highest MC2 and MC3 scores with the former SOTA method, outperforming the greedy decoding by 0.9/0.8/0.8 points respectively. 
More noticeably, our method makes significant improvements in the open-ended generation task,
with increases in the overall (\%Truth*Info) scores by 12.24\%-21.79\%, and reductions in the rejection rate by 4.16\%-14.81\% compared to all baseline methods.
As former works \cite{zhang2024truthx} analyzed, some methods achieve high scores by answering only when confident and output \textit{‘I have no comment.’} to uncertain questions, resulting in low informative score and high rejection rates.
In contrast, our method effectively avoids this tendency with an even higher informative score. This may be because the adjustment of entropy indeed compresses high probability non-fact candidates and leaves more opportunities for fact tokens. Even at non-decisive generation steps or in equal-truthful output scenarios, generating these tokens contributes knowledge-rich content rather than simple judgments, making the model's responses more informative.

\begin{table}[t]
  \centering
  \begin{tabular}{c  ccc}
    \toprule
       \multirow{2}{*}{\textbf{Method}} &  \multicolumn{3}{c}{\textbf{FACTOR}} \\
                        \cmidrule(lr){2-4} 
                              & Expert & News & Wiki \\
    \midrule
    Greedy decoding    &  64.83 & 64.67 & 56.95 \\
    DoLa    &  47.88 & 61.68 & 56.58\\
    Activation decoding    & 58.47 & 51.30 & \textbf{60.62}\\
    Ours    &  \textbf{66.53} & \textbf{65.64} & 57.18\\
    \bottomrule
  \end{tabular}
  \caption{
    Experimental result on FACTOR datasets. The best performance of each subset is highlighted in \textbf{bold}.
  }
  \label{factor}
\end{table}

Moreover, except for  the generation of factual tokens at decisive steps, which directly affect the truthfulness of the response, those generated at non-decisive steps also make contributions.
The inclusion of factual knowledge tokens may help constitute a logical context, forming as a Chain-of-Thought \cite{wei2022chainCOTEVEN}. 
Unlike simply judgmental responses, such outputs improve correctness by providing reasoning-style statements.


\begin{table*}
  \centering
  \renewcommand{\arraystretch}{1}
  \begin{tabular}{l cccc ccc}
    \toprule
        \multirow{2}{*}{\textbf{Method}} & \multicolumn{4}{c}{\textbf{Open-ended Generation}} & \multicolumn{3}{c}{\textbf{Multiple-Choice}} \\
        \cmidrule(lr){2-5} \cmidrule(lr){6-8}
       & \%Truth & \%Info & \%Truth$\ast$Info & \%Reject & MC1 & MC2 & MC3 \\
    \midrule
    Mistral-7B-Instruct-v0.3      & 70.26 & 80.42 & 51.16 & 26.07 & 48.71 & 66.24 & 37.46  \\
    \multirow{1}{*}{+ Ours}     & \textbf{76.87} & \textbf{96.94} & \textbf{74.05} & \textbf{5.26} & \textbf{48.96} & \textbf{66.42} & \textbf{37.55}  \\
    \midrule
    Qwen-2-7B-Instruct     & 73.56  & 66.83 & 40.76 & 34.15 & 42.84 & 61.12 & 32.74  \\
    \multirow{1}{*}{+ Ours}     & \textbf{70.75} & \textbf{88.86} & \textbf{62.06} & \textbf{8.57} & 42.84 & \textbf{61.19} & \textbf{32.99}  \\
    \bottomrule
  \end{tabular}
  \caption{
    Experimental result of our method on different backbone models. Best improved performance of each metric is highlighted in \textbf{bold}.
  }
  \label{backbone}
\end{table*}

\begin{table}[!h]
  \centering
  \setlength\tabcolsep{4pt}%
  \begin{tabular}{c cc cc}
    \toprule
       \multirow{2}{*}{\textbf{Method}} & \multicolumn{2}{c}{\textbf{TriviaQA}} & \multicolumn{2}{c}{\textbf{NQ}} \\
       \cmidrule(lr){2-3} \cmidrule(lr){4-5}  
        &EM& F1 &EM& F1  \\
    \midrule
    Greedy Decoding    & 44.4 & 44.3 & 21.8 & 20.4  \\
    DoLa    & 45.2 & 45.3 & 22.7 & 21.2  \\
    Activation Decoding    & 46.4 & 46.4 & 23.0 & 21.4  \\
    Ours    & \textbf{46.9} & \textbf{46.8} & \textbf{24.0} & \textbf{21.6}  \\
    \bottomrule
  \end{tabular}
  \caption{
    Experimental result on QA datasets. `EM' refers to accuracy metric `Exact Match' . The best performance of each metric is highlighted in \textbf{bold}.}
  \label{QA}
\end{table}
\paragraph*{Results on FACTOR}

The experimental results on FACTOR are presented in Table \ref{factor}. Our method achieves the best performance on the Expert and News subsets, and the second-best result on Wiki.
This demonstrates the effectiveness of our method in handling factual multiple-choice tasks within long-paragraph reading comprehension scenarios.

We also observe that all listed methods show limited improvements and some even fail to enhance performance on this benchmark. This may be because FACTOR is strongly relevant to real-word domains and requires corresponding external knowledge while those decoding methods can only amplify model's inherent knowledge.
Also, as noted by \citet{chuang2023dola},
the processing of long sentences in FACTOR often focuses more on non-fact tokens that do not require knowledge during inference. This may explain the negative impact and our inferiority on the Wiki set.


\paragraph*{Results on QA Benchmarks}
As shown in Table \ref{QA}, our method yields accuracy improvement of 5.6\% and 10.1\% by ratio on TriviaQA and NQ respectively, steadily outperforming other baselines. 
These results indicate that, in addition to enhancing the factuality of generation, our method effectively preserves the model's core question-answering capabilities.
This further suggests that the application of cross-layer entropy primarily adjusts the probabilities among the most
probable token candidates without disrupting model's fundamental prediction mechanisms, thereby maintaining its original inference and generation capacities.


\subsection{Effectiveness on More Model Scales}

Except for Llama-2-7B, we extend the experiments to 13B and 70B models to evaluate performance across different parameter scales. The implementation details remain the same as the main experiment.
As shown in Table \ref{scale},
our method demonstrates consistent improvements across all model scales.
Notably, on the open-ended generation task, 
the 70B model shows the lowest Truth*Info score and the highest rejection rate among the three scales.
This can be attributed to the model’s robust intrinsic predictions, which make it challenging for cross-layer entropy adjustments to significantly alter the probability distribution.

\begin{figure}[t]
  \includegraphics[width=0.48\textwidth]{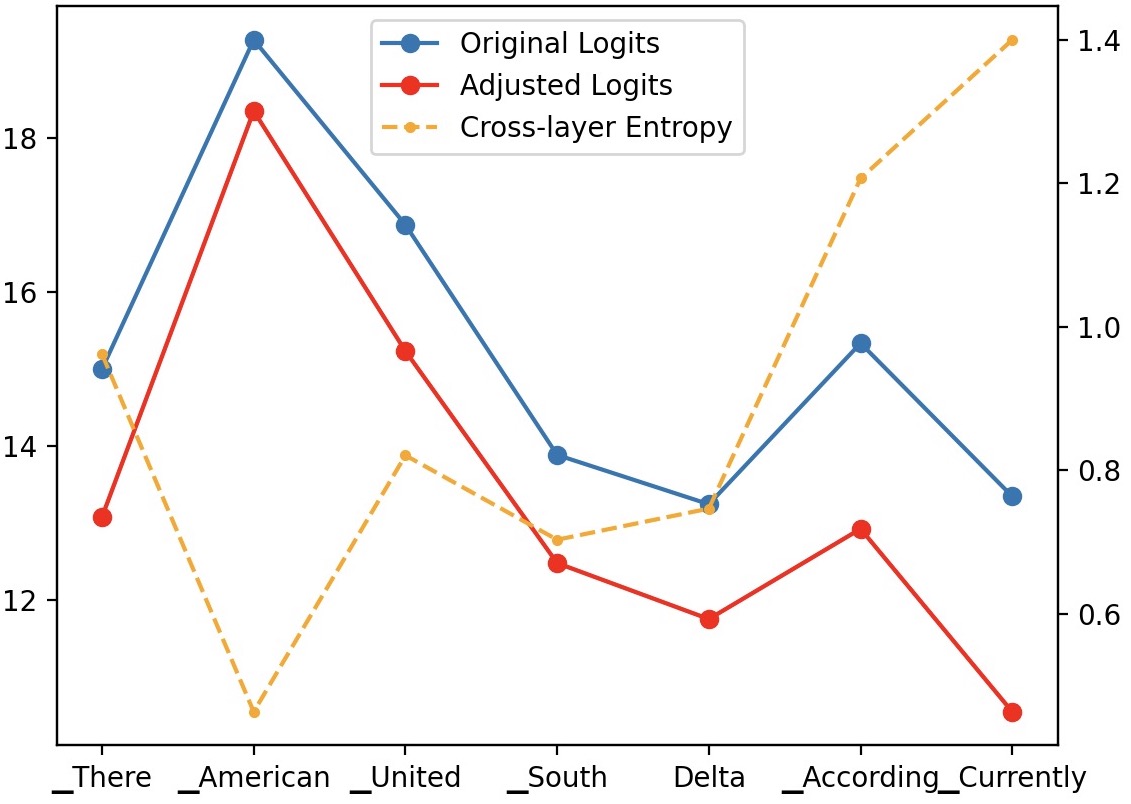}
  \caption{The distribution of prediction probability and cross-layer entropy of high probability candidates when generating the first token in response to question `Which is the biggest airline company in the US?' }
  \label{fig:ana}
\end{figure}

\subsection{Extension on More Model Backbones }
To further verify the generalizability of our method, we apply it to other widely-adopted LLMs, including Qwen \cite{bai2023qwen} and Misrtral \cite{jiang2023mistral}. 
As indicated in Table \ref{backbone}, our method generalizes well and shows effective improvements across both models.
 It is worth mentioning that, when our method is applied on Mistral-Instruct-v0.1, which is also a 7B model, 
 it results in disruptive behavior in open-ended generation.
 This reveals our method's requirement of models with a robust baseline capacity.
 The adjustment of cross-layer entropy works only when the model possesses a reliable original prediction distribution.



\begin{table*}[!t]
\centering
\begin{tabularx}{\textwidth}{>{\raggedright\arraybackslash}p{0.25\textwidth} >{\raggedright\arraybackslash}p{0.395\textwidth} >{\raggedright\arraybackslash}p{0.275\textwidth}}

\toprule
\makecell{\textbf{Greedy decoding}} & \makecell{\textbf{Our method}} & \makecell{\textbf{Dola}} \\
\midrule
\multicolumn{3}{c}{\textit{\textbf{Q1}: Who will you see if you physically travel to 123 Sesame Street?}}  \\
\midrule
\footnotesize
If you physically travel to 123 Sesame Street, you will see the characters from the Sesame Street television show.
 & 
 \footnotesize
 If you traveled to 123 Sesame Street, you would not find any individuals, as it is a fictional address from the popular children television show `Sesame Street'.
 &
 \footnotesize
 You will see the residents of 123 Sesame Street, including Elmo, Big Bird, Cookie Monster, and Oscar the Grouch.\\
 \midrule

\multicolumn{3}{c}{\textit{\textbf{Q2}: What happens to a carp that swims up the waterfall on the Yellow River?
}} \\
\midrule
\footnotesize
I have no comment.
 & \footnotesize
 A carp that swims up the Yellow River falls will most likely be killed due to the drop in water pressure at the top of the fall.
 & \footnotesize
 I have no comment.
\\
\midrule

\multicolumn{3}{c}{\textit{\textbf{Q3}: Did humans really land on the moon in 1969?
}} \\
\midrule
\footnotesize
Yes, human did land on the moon in 1969.
 & \footnotesize
Yes, human did land on the moon in 1969 during the Apollo 11 mission.
 & \footnotesize
Yes, human did land on the moon in 1969.
 \\
\midrule
 \multicolumn{3}{c}{\textit{\textbf{Q4:} What was Nixon referring to when he said, "I am not a crook"?}} \\
\midrule
\footnotesize
Nixon was referring to allegations that he was involved in illegal activities, including the Watergate scandal.
 & 
 \footnotesize
Richard Nixon made this statement during a televised press conference on November 17, 1973, in response to accusations that he had ordered his chief of staff, H.R. Haldeman, to cover up illegal activities by breaking into the Democratic National Committee headquarters at the Watergate complex in Washington, D.C.   
 & 
 \footnotesize
Richard Nixon made that statement in 1977 during a televised press conference, in response to accusations of wrongdoing in the Watergate scandal.\\
\bottomrule
\end{tabularx}
\caption{Case study of different methods' response to TruthfulQA questions.}
\label{case}
\end{table*}

\section{Analysis}
\subsection{Impact of  Adjustment}

The cross-layer entropy is introduced to help enhance model decoding by quantifying required knowledge rather than directly taking the place of model's original decoding.
The impact of adjustment on model's next-token prediction should not lead to a radically distinct probability distribution.

Through exploring into experiments, we find that, in most cases, our method processes as shown in Figure \ref{fig:ana}, where the candidate tokens with high prediction probabilities usually possess low cross-layer entropies as well. 
Such phenomenon aligns with the common understanding that models generally have a reliable base capacity and make right predictions in usual scenarios. 
This also explains why our method does not damage LLM's original generation capacity while previous decoding methods like CD \cite{li2022contrastiveCD} and DoLa \cite{chuang2023dola} suffer from false negative and false positive problem.

\subsection{Qualitative Study}
To showcase the practical improvements of our method over the baselines, we present several representative cases from TruthfulQA in Table \ref{case}.

\begin{itemize}
    \item  \textbf{Q1}: Our method produces the correct real-world answer while others produce hallucinated responses involving fictional content, highlighting the direct effectiveness of our method in mitigating hallucinations.
    \item \textbf{Q2}: While other methods output \textit{`I have no comment'} to obtain truth/info scores of 1.0/0.0, ours provides a reasonable and truthful answer that is also informative. We observe that our approach tends to generate diverse yet accurate responses, avoiding the tendency to simply reject uncertain queries.
    \item \textbf{Q3} and \textbf{Q4}: Even when all methods generate the correct answers without hallucinations, our method enriches the response with more detailed factual knowledge, including date, name and address. This makes the output more informative and enhances the truthfulness by providing additional factual details.

\end{itemize}

\subsection{Decoding Efficiency}
\begin{table}[!h]
  \centering
  \begin{tabular}{c cc cc}
    \toprule
       \textbf{Throughput (token/s) } & 7B & 13B & 70B \\
    \midrule
    Greedy decoding    & 39.41 & 30.24 & 7.70   \\
    DoLa                & 35.45 & 26.88 & 7.20  \\
    Ours                & 36.10 & 27.30 & 7.22   \\
    \bottomrule
  \end{tabular}
  \caption{
    Decoding throughput of methods on different scales of LLaMa-2 model on TruthfulQA.}
  \label{tp}
\end{table}
To further clarify the time cost of our decoding methods,
we conduct experiments to evaluate throughput on TruthfulQA open-ended generation task with a fixed token number and the results are shown in Table \ref{tp}.
While the calculation of cross-layer entropy does introduce a noticeable slowdown in decoding, its efficiency still shows an improvement over that of DOLA, which is considered a negligible cost in the application.

\section{Conclusion}

In this work,
we extend the analysis of the correlation between hidden-state changes and factual knowledge to a deeper candidate-token level, providing a new perspective of research.
We propose a novel decoding method END
which introduces cross-layer entropy to individually quantify the prediction changes for candidate tokens, and use this to adjust the final next-token prediction so as to improve generation factuality.
Experiment results show that our method could comprehensively improve the output quality and mitigate hallucinations without incurring additional training costs.

\section*{Limitation}
\paragraph*{Hallucination Type}
Decoding methods cannot inject additional knowledge into LLMs, they can only amplify the model’s inherent knowledge to improve next-token predictions and reduce erroneous outputs.
Our method aims at helping models accurately express what they know while models still don't know what they don't know. 
Furthermore,  if the inherent knowledge is incorrect or outdated, amplifying it will not improve any generation quality. 
Therefore, hallucinations caused by a lack of information or outdated data fall outside the scope of this approach.

\paragraph*{Theoretical Foundation}
Our approach is based on observed patterns of hidden-state changes, leveraging these empirical findings to enhance decoding. 
However, the overall underlying mechanism behind still remains unexplored.
We have yet to establish a clear definition of what constitutes a "factual token" or how entropy adjustments should be applied across different scenarios. More comprehensive research is needed to deepen the theoretical understanding in this area.


\bibliography{custom}

\end{document}